\documentclass[10pt]{article} 
\usepackage[accepted]   {rlc}

\usepackage{amssymb}            
\usepackage{mathtools}          
\usepackage{mathrsfs}           
\mathtoolsset{showonlyrefs}     
\usepackage{graphicx}           
\usepackage{subcaption}         
\usepackage[space]{grffile}     
\usepackage{url}                

\usepackage{tikz}

\title{Value Internalization: Learning and Generalizing from Social Reward}

\author{{\large \bf Frieda Rong } \\
  Stanford University \\ rongf@cs.stanford.edu
  \And {\large \bf Max Kleiman-Weiner } \\
 University of Washington\\ maxkw@uw.edu}

\begin{document}

\maketitle

\begin{abstract}
Social rewards shape human behavior. During development, a caregiver guides a learner’s behavior towards culturally aligned goals and values. How do these behaviors persist and generalize when the caregiver is no longer present, and the learner must continue autonomously? Here, we propose a model of value internalization where social feedback trains an internal social reward (ISR) model that generates internal rewards when social rewards are unavailable. Through empirical simulations, we show that an ISR model prevents agents from unlearning socialized behaviors and enables generalization in out-of-distribution tasks. We characterize the implications of incomplete internalization, akin to ``reward hacking'' on the ISR. Additionally, we show that our model internalizes prosocial behavior in a multi-agent environment. Our work provides a foundation for understanding how humans acquire and generalize values and offers insights for aligning AI with human values.
\end{abstract}

\section{Introduction}
Why do we want what we want? Some goals we pursue are responses to the extrinsic rewards and punishments of the environment. We pursue food when hungry, shelter when cold, and sleep when tired. Money can motivate us to work harder, and the threat of punishment can incentivize us to follow the law.  Other goals are intrinsically self-motivated and do not require external reinforcement. We play and explore, feel a warm glow when altruistic, and may take pride in our work even when no one is watching \citep{andreoni1990impure, ryan2000intrinsic}. Both extrinsic and intrinsic rewards have likely been shaped by natural selection to enable adaptive behavior across many environments \citep{singh2009rewards}. They have also played a key role in building reinforcement learning agents that can learn in an open-ended fashion across a lifetime of experiences and tasks without hand-crafting reward functions for each one  \citep{singh2010intrinsically, schmidhuber2010formal, mohamed2015variational, kulkarni2016hierarchical, jaques2019social}. While some pursue a quest for a universal reward function that generates the full suite of human-like intelligent behavior \citep{silver2021reward}, we aim to study how values might be acquired through social and cultural learning and then leveraged for open-ended autonomy. 


Our approach can explain some key challenges for understanding the source of values. First, although many aspects of desire are innate, and any acquisition process itself requires some degree of innate motivation and machinery, there must be a substantial role for learning in determining what humans find rewarding. Different cultures across time and space have varied substantially in terms of what people in those societies find rewarding \citep{henrich2001search, henrich2006costly, medvedev2024motivating}. In some places, spicy food can cause physical pain, while in others, food without spice is considered bland and tasteless \citep{billing1998antimicrobial}. Different individuals chase meaning and reward in different ways: maximizing money, power, artistic expression, knowledge, fame, the probability of reaching an afterlife, and many others \citep{maslow1958dynamic}. Moral values, such as how different individuals trade off the welfare of different groups, vary as well; some might weight family members highly, while others strive for impartiality \citep{kleiman2017learning, mcmanus2020we}. This logic applies to even more basic aspects of daily thinking. Is curiosity a virtue to be celebrated and inculcated in children, or a vice (``curiosity kills the cat''), and is it best repressed or inhibited? This broad diversity suggests that a satisfactory explanation will have learning play a key role. 

Second, while different environments might differentially shape what one finds rewarding to some extent, it is unlikely that differences in the physical environment alone are sufficient to fully explain human variation. 
Most environments are highly open-ended, where correct behavior cannot be reduced to a single goal specification or clear metric for success \citep{stanley2015greatness}. Outside of the most basic needs, such as survival, the importance of a given goal is often determined collectively and specific to one's culture. Even within the narrow context of a video game, there are many ways to play: go for the highest score, ``speedrun'' to finish the game as fast as possible, explore every nook and cranny, find exploits, create games within the game, and more. 

We address these two challenges by proposing that to the extent environments have relevant rewards or reward-relevant information; those rewards often come from social influences \citep{bandura1963influence, ho2017social, magid2017moral}. 
The structure of this information can take many forms. Direct forms of feedback, such as praise, smiles, laughs, punishments, comparison, and correction, and more indirect forms of feedback, such as instruction or demonstration \citep{jeon2020reward}. Children's interactions with their caretakers are rich in this kind of feedback, shaping human reward learning from an early age \citep{grusec1994impact}. Yet learning from social rewards contains a computational puzzle. If the source of reward is social, it will not be available when the social partner isn't present. From a developmental perspective, while a caregiver might provide a learning signal early on -- ultimately, the learner will need to continue their learning, exploration, and autonomy without supervision. This is a problem for any system that learns from reinforcement -- if rewards disappear from an environment, the behavior those rewards incentivized will quickly be extinguished. Clearly, this does not happen for human learners. 

Here, we propose that learners sustain exploration and autonomy when social reward subsides by \textit{internalizing} their caregiver's rewards. This requires the ability to model the caregiver's rewards in a way that generalizes to the new environments the learner faces. This idea is prominent in attachment theory and is called an internal working model \citep{bowlby1969attachment, ainsworth1978patterns, johnson2007evidence}. In our work, we develop a novel paradigm for studying the challenge of generalizing from social rewards. First, we extend the Markov Decision Process (MDP) formalism so that environmental rewards are augmented with social feedback that is only present temporarily. Second, using a suite of navigation tasks developed with this framework, we demonstrate the abovementioned challenge and show that a baseline reinforcement learning (RL) agent unlearns their goal-directed behavior once social rewards are removed. Third, we develop an RL agent that internalizes the rewards of others and show that it solves this key challenge. Finally, we test this agent in a variety of different challenges and study its limitations in generalizing both within the training distribution and to new more demanding tasks, internalizing self and prosocial rewards, and overcoming reward hacking. Together, this work proposes a framework for analyzing value internalization, formalizes the key challenges, and proposes a new agent that addresses these challenges and captures aspects of human value internalization. 

\subsection{Related Computational Work}
Our work takes inspiration from reinforcement learning from human feedback (RLHF), a technique currently used to train agents and align models to judgments made by human annotators. \citet{christiano2017deep} train a reward model from pairwise preference judgments of an agent's behavior and show that the reward model can be used to train a deep reinforcement learning agent on simple tasks. \citet{tien2022causal} study generalization in reward models and show that reward modeling from pairwise judgment data often fails to generalize because the reward models can learn spurious correlations rather than capturing the underlying causal process. Similar to our work here, \citet{colas2020language} train a goal generator from the language of a social partner and show that this goal generator can imagine new goals to improve generalization and exploration. Finally, \citet{kleiman2017learning} develop a hierarchical probabilistic model for the moral domain that learns to set the weights of a multi-attribute utility function depending on the observations made by the learner.

\begin{figure}[!t]
    \begin{minipage}[b]{.5\linewidth}
    \begin{subfigure}[b]{.18\linewidth}
    \includegraphics[width=\textwidth]{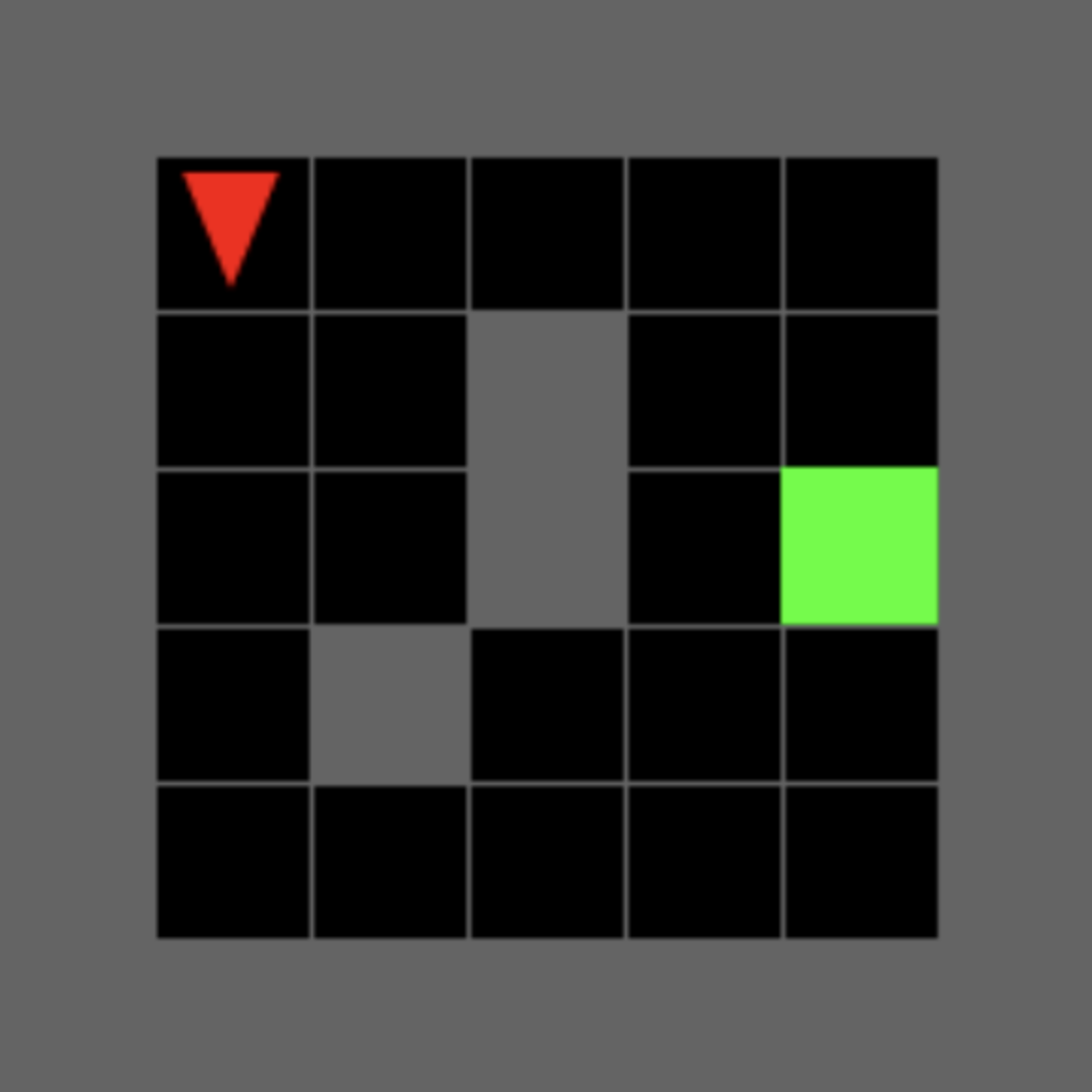}\\
    \includegraphics[width=\textwidth]{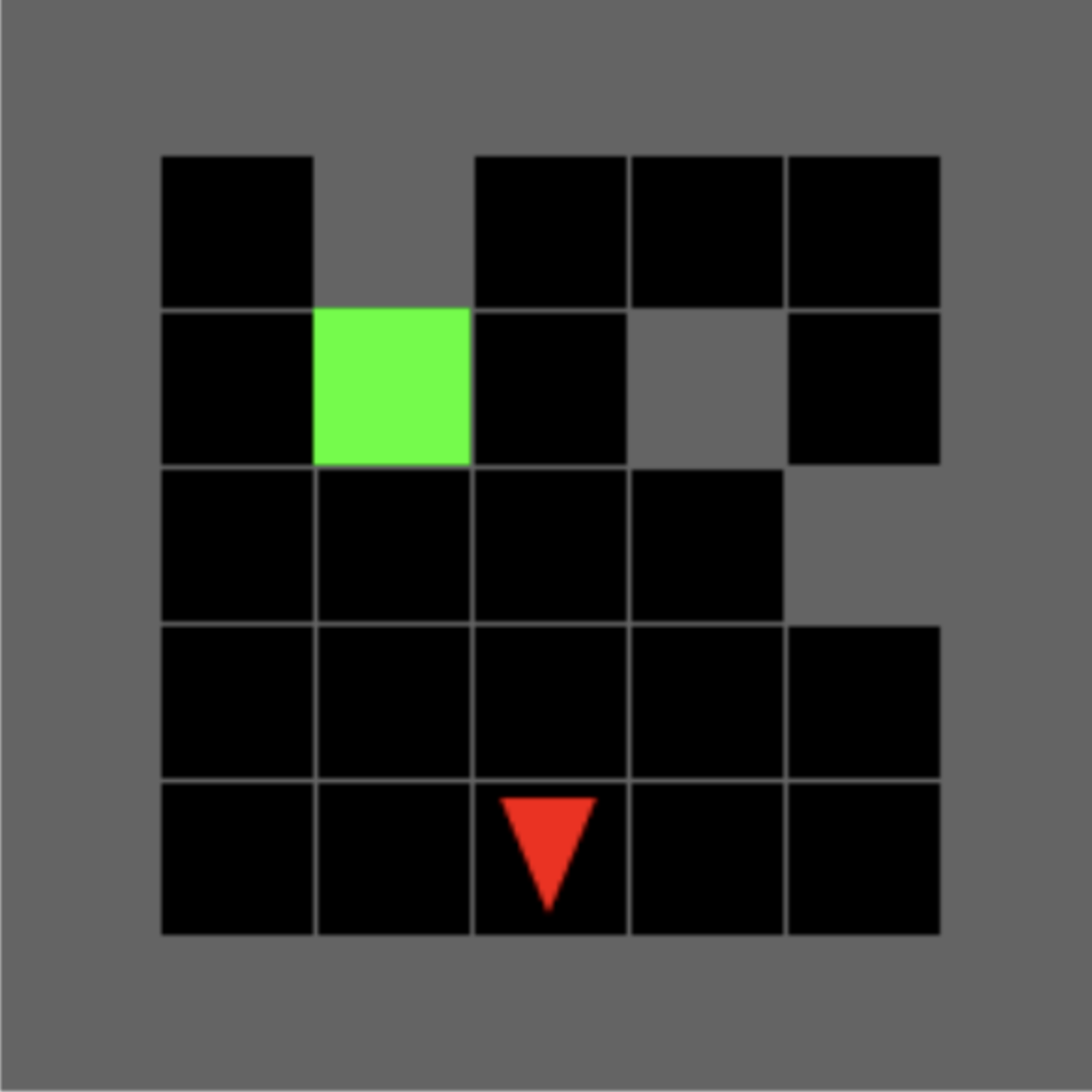}\\
    \includegraphics[width=\textwidth]{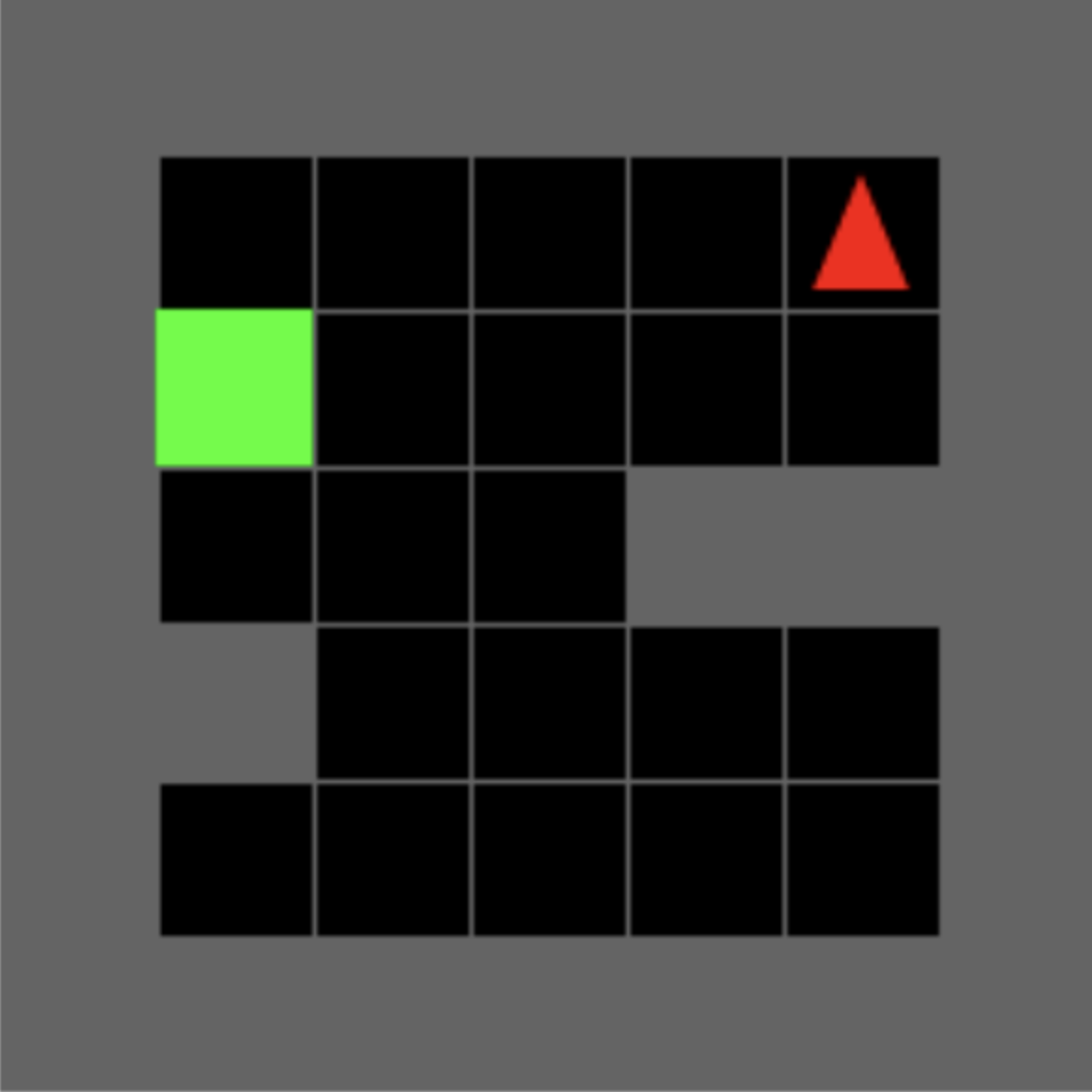}
    \label{fig:grid}
    \end{subfigure}
    \hfill
    \begin{subfigure}[b]{0.72\linewidth}
    \includegraphics[width=\textwidth]{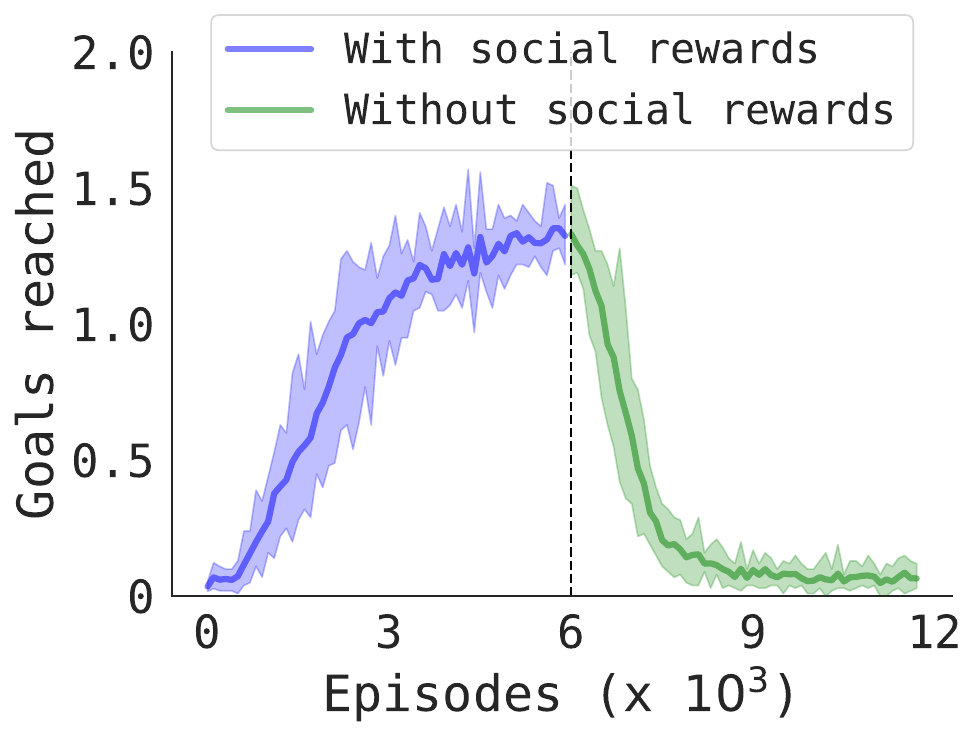}
    \label{fig:challenge_graph}
    \end{subfigure} 
    \end{minipage}
    \hfill
    \begin{minipage}[b]{.5\textwidth}
    \caption{The challenge of learning from social rewards. (left) Three example grids from our environment. The goal square is shown in green, and the agent is the red triangle. Three obstacles shown in grey are randomly arranged in each grid.
    (right) Learning with (blue) and without (green) social rewards. A baseline reinforcement learning agent learns to navigate to the green square when the caregiver is present. The goal-directed behavior is unlearned once the caregiver leaves (dotted vertical line at 6K episodes for the green trace). Traces averaged over ten seeds and smoothed. Bands show the min and max.}
    \label{fig:challenge}
    \end{minipage}

\end{figure}

\section{MDPs With Social Rewards}

We study the process of value internalization in a two-agent Markov decision process (MDP) with a learner and a caregiver. In our setup, the caregiver only interacts with the learner by giving social rewards. Social reward is a single continuous number corresponding to the degree to which the feedback is intended to be rewarding (positive) or punishing (negative). Finally, in some trials, the caregiver is absent, so there is no social reward in those trials. 

Formally, an MDP with social rewards (MDP-SR) is a tuple $\langle \mathcal{S}, \mathcal{A}, \mathcal{T}, \gamma, \mathcal{R}_e, \mathcal{P}, \mathcal{R}_s \rangle$: a set of states $\mathcal{S}$, a set of actions for each state $\mathcal{A}(s)$, a transition function that maps states and actions to future states $\mathcal{T}(s, a) \rightarrow s'$, 
a discount factor $\gamma \in [0,1)$, an extrinsic reward function that maps actions and outcomes to environmentally given rewards $\mathcal{R}(s, a, s')_e \rightarrow \mathbb{R}$. We extend these terms to account for social reward by augmenting the MDP with $\mathcal{P}(s) \in \{0,1\}$ that indicates whether the caregiver is present (1) or absent (0) and the social reward $\mathcal{R}_s(s, a, s') \rightarrow \mathbb{R}$ which is available only when $P(s)=1$. We assume that learners are socially motivated and have a utility function $U = R_e + R_s$ that integrates environmental and social rewards \citep{dweck2017needs}. The learner aims to find a policy $\pi$ that maximizes expected cumulative discounted utility. 

Our experiments are divided into two phases. First is a socialization phase where the caregiver is present ($p=1$). Second is an autonomous phase where the caregiver is absent ($p=0$). Our framework allows for more complex dynamics (e.g., slowly reducing the probability of the caregiver's presence over time), but we use a simple two-phase approach to simplify the analyses. The MDP-SR framework enables us to ask questions about how computational learners will handle the transition between these two phases. 

\begin{figure}[!b]
    \begin{subfigure}[b]{.5\linewidth}
    \includegraphics[width=\linewidth,trim={0 7cm 0 0},clip]{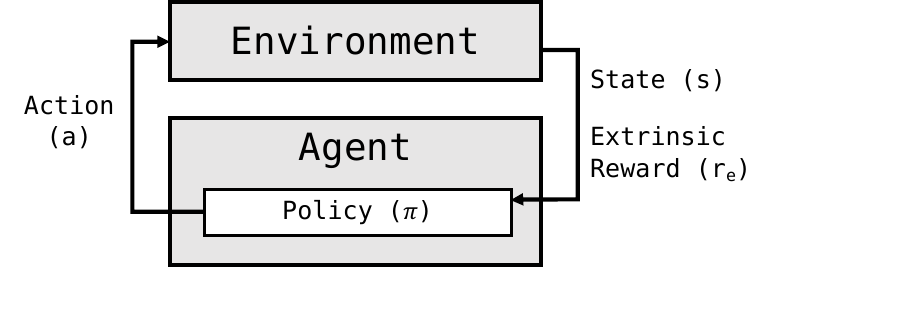}
    \end{subfigure}
    \begin{subfigure}[b]{.5\linewidth}
    \includegraphics[width=\linewidth,trim={0 0 0 5cm},clip]{CameraReady/LaTeX/fig/rl.pdf}
    \end{subfigure}
    \caption{Agent architectures. (left) Standard view of reinforcement learning with extrinsic reward from the environment. (right) Learning from social rewards. Dotted lines indicate that the caregiver and the social rewards they give are not always present. When present, social rewards affect the policy as well as train an internalized social reward model (ISR) that provides internal rewards when the caregiver is absent. }
    \label{fig:rl_loop}
\end{figure}

We developed a procedurally generated set of navigation tasks using the Minigrid Learning Environment \citep{MinigridMiniworld23}. Figure~\ref{fig:challenge} shows some examples. In each episode, we generate a 5x5 grid with the agent denoted as a red arrow that can face any of the cardinal directions, a green square, and three blocks that create obstacles for navigation. The green square, three blocks, starting position, and agent orientation are uniformly randomly sampled. The agent can turn 90 degrees in place or move forward one square. Going forward has a small negative cost, $R_e = -\frac{0.9}{\max(\text{steps})}$, where $\max(\text{steps})$ is the maximum number of steps in an episode. There were 20 steps in each episode, so $R_e = -0.045$. This small cost incentivizes efficient action and is the only extrinsic reward in our setting. The discount rate $\gamma=0.99$. The grid and starting location are randomly resampled if the agent reaches the green square. During the first phase (socialization), when the caregiver is present, the caregiver provides a large reward ($R_s = 0.4$) when the agent reaches the green square. 

\section{Modeling Value Internalization}
Our baseline agent is a deep reinforcement learner trained with PPO from Stable Baselines 3 \citep{schulman2017proximal, raffin2021stable}. We use a learning rate of 1e-4 and otherwise use the default hyperparameters from Stable Baselines 3 for all models tested. 
The top of Figure~\ref{fig:rl_loop} shows an abstracted version of the typical loop between the environment and the agent where the environment provides the state and an extrinsic reward, and the agent produces actions based on its learned policy. 

Figure~\ref{fig:challenge} shows the performance of the baseline agent on our environment distribution. We contrast what happens when the caregiver is present (blue) versus when the caregiver leaves at the halfway point (green). When the caregiver remains, the agent steadily improves its performance until eventually plateauing near an optimal level. In contrast, when the caregiver leaves at the halfway point, performance rapidly drops to zero. This confirms our initial hypothesis: when the social rewards provided by the caregiver are the primary source of rewards that define the task, a typical reinforcement learner will not be able to continue learning and exploring autonomously when the caregiver is no longer present.

We hypothesize that human learners address this problem by internalizing the values of others. Here, we formalize this hypothesis by augmenting our baseline agent with an internalized social reward model (ISR) that learns to model the social rewards given by the caregiver and creates internal rewards ($\mathcal{R}_i$) when the caregiver is absent (Figure~\ref{fig:rl_loop}). Thus we can write the full utility function of an agent with an ISR as $U=R_e + P\cdot R_s + (1-P)\cdot R_i$. It receives a non-zero social reward $R_s$ when the caregiver is present $P=1$ and a non-zero internalized reward $R_i$ when the caregiver is absent $P=0$. 

The ISR model is a deep neural network using the same architecture as the policy network. The network takes in the state and action and predicts reward. During the socialization phase, the agent stores the social rewards received, and those stored rewards are used to train the ISR model. The model was trained to minimize mean square error (MSE, $||R_s-R_i||$) since rewards are continuous. Finally, since the distribution of social rewards is imbalanced -- positive rewards are more sparse than zero rewards -- rewards were sampled such that each training batch had a balanced sample of reward magnitudes.

When the task or distribution of tasks changes, deep RL policies often fail to generalize \citep{kansky2017schema}. This failure results partly from the challenge of needing to predict an entire sequence of actions that optimize the expected cumulative discounted rewards. In contrast, the ISR module only needs to predict the reward for a particular action in a particular state without considering future actions. If the ISR module generalizes to new environments before the policy does, the agent could continue learning in those new environments even in the total absence of extrinsic reward. On the flip side, if the ISR fails to generalize, then the agent will learn to optimize a misspecified reward \citep{pan2022effects, tien2022causal}. This could lead to reward hacking where the agent successfully optimizes its reward signal, but that reward no longer matches what the caregiver intended \citep{skalse2022defining}. We study these possibilities empirically in the next section. 

\begin{figure}[!tb]
    \begin{minipage}[b]{.58\linewidth}
    \begin{subfigure}[b]{.48\linewidth}
        \includegraphics[width=\textwidth]{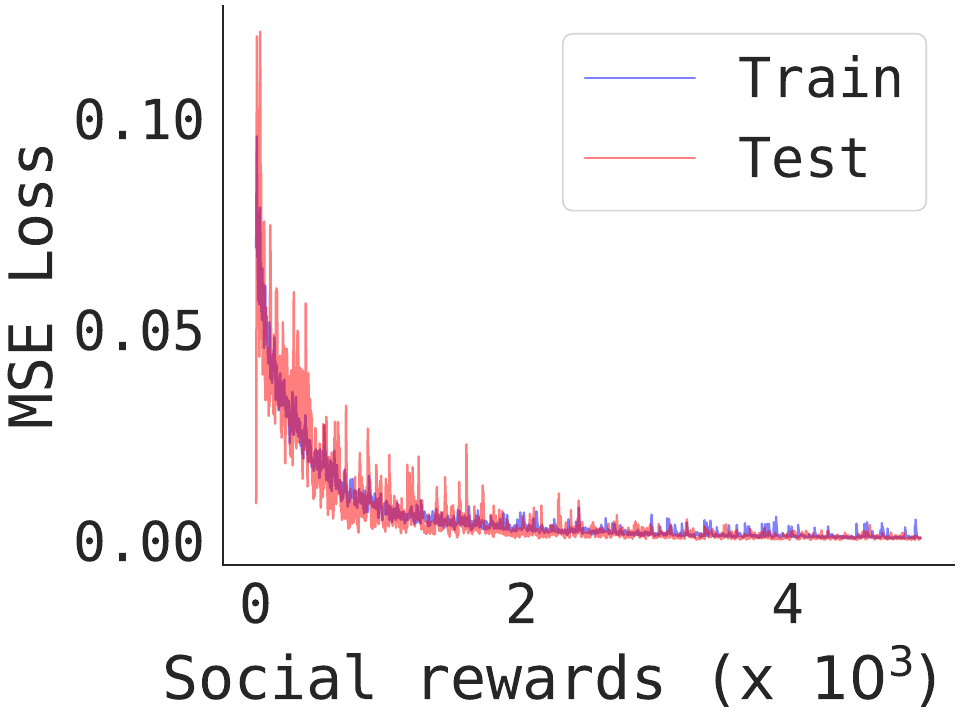}        
    \end{subfigure}
    \hfill
    \begin{subfigure}[b]{.48\linewidth}
        \includegraphics[width=\textwidth]{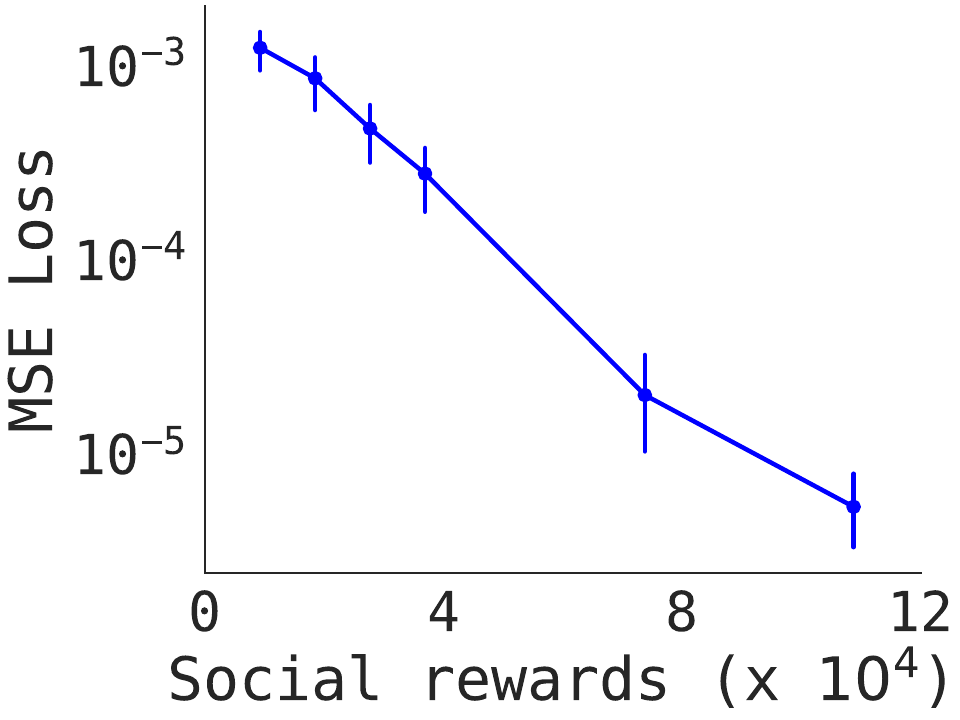}
    \end{subfigure}
    \end{minipage}
    \hfill
    \begin{minipage}[b]{.4\linewidth}
    \caption{Training the internalized social reward (ISR) model. (left) Example training curve for the ISR model trained on social rewards. The model quickly converges with no measurable gap between train and test performance. (right) ISR test loss continually decreases when trained with more social rewards. Results averaged over ten seeds and smoothed. Error bars are the standard error.}
    \label{fig:isr_train}
    \end{minipage}
\end{figure}

\section{Results}
We first analyze the training of the ISR model. We then test whether a reinforcement learner augmented with ISR can solve the challenge posed in Figure~\ref{fig:challenge} and analyze whether the ISR enables generalization by allowing for additional learning even without any social reward. Finally, we introduce a multi-agent scenario where the caregiver rewards altruistic behavior and show that our model extends to prosocial value internalization.

\subsection{Training the ISR model} 
Figure~\ref{fig:isr_train} shows learning curves for the ISR. With sufficient data, the model achieves minimal test loss. The final test loss was an exponential function of the amount of social rewards observed, where each doubling of the number of rewards yielded an order of magnitude reduction in MSE loss. 

\subsection{Continual Learning and Generalization}
\begin{figure}[!t]
    \centering
    \begin{minipage}[b]{0.6\linewidth}
    \begin{subfigure}[b]{.5\linewidth}
    \includegraphics[width=\textwidth]{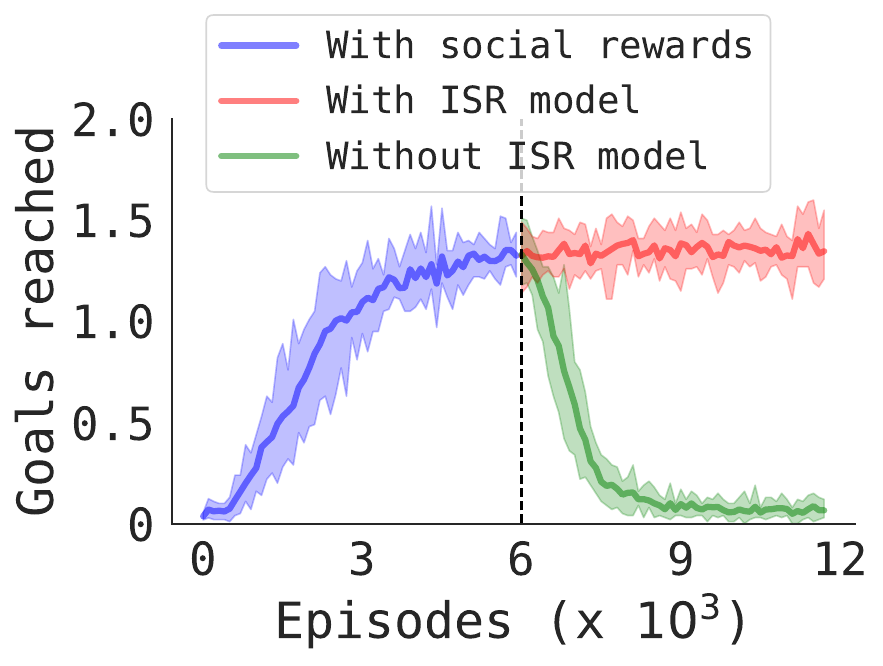}        
    \end{subfigure}%
    \hfill
    \begin{subfigure}[b]{.4\linewidth}
    \hfill\includegraphics[width=.35\textwidth]{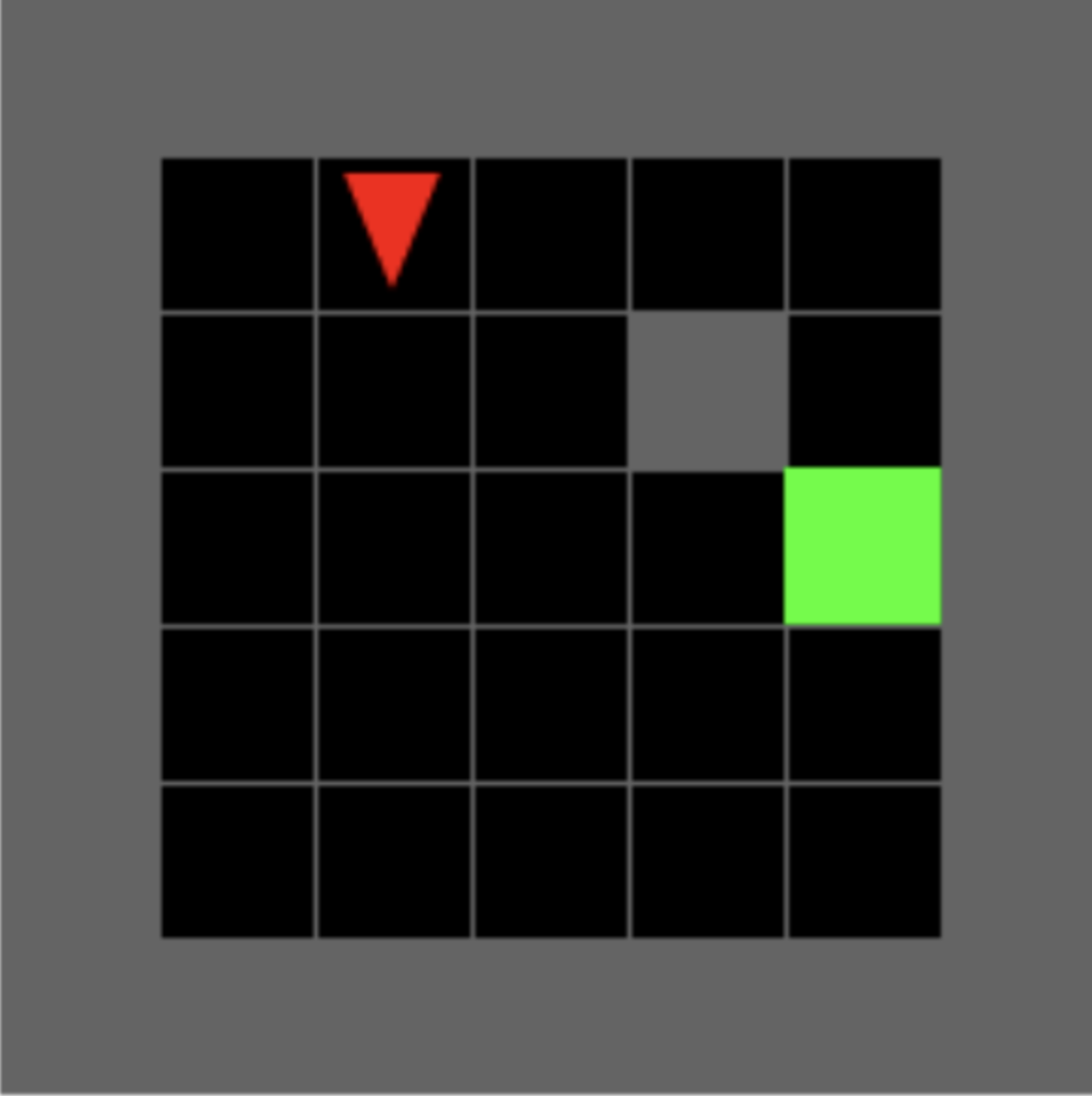}
    \raisebox{1.4em}{$\rightarrow$}
    \includegraphics[width=.35\textwidth]{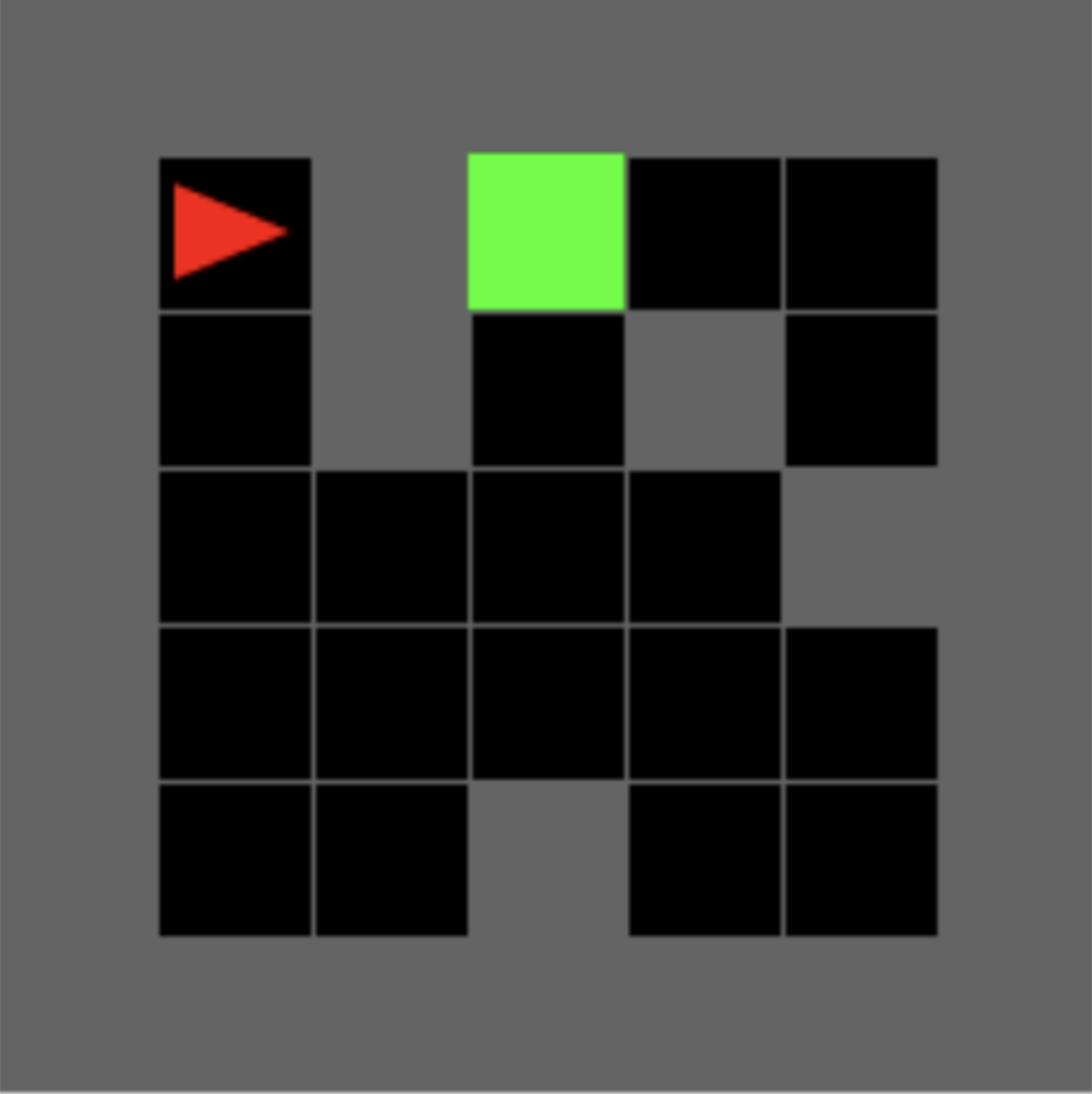}\hfill   
    \includegraphics[width=\textwidth]{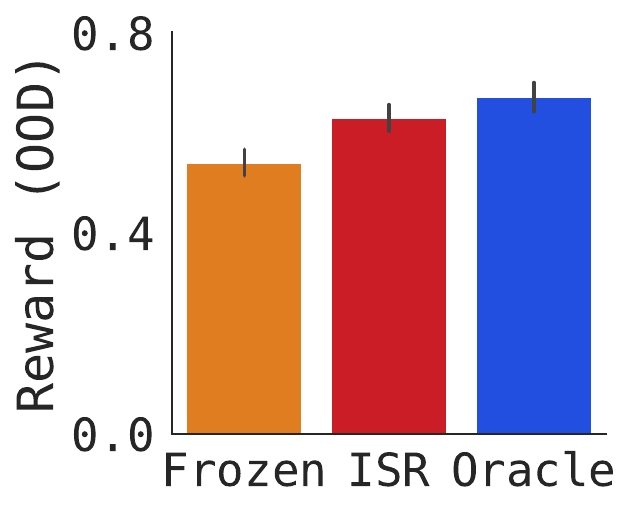}
    \end{subfigure}
    \end{minipage}\hfill%
    \caption{The ISR model prevents unlearning and enables out-of-distribution (OOD) generalization (left) Agents first learn with social rewards from the caregiver (blue). After 6K episodes, the caregiver is removed (vertical dotted line). Without the ISR model, the agent quickly unlearns the behavior (green). The ISR model prevents unlearning with no measurable loss in performance (red). Results averaged over ten seeds and smoothed. Bands show the min and max. (right) Comparing OOD generalization where models were trained with one block and must generalize to five. The ISR performance was significantly greater than the frozen baseline ($p<0.05$, t-test) and not significantly different from the oracle ($p=0.32$, t-test). See text for model descriptions. 
    Results averaged across ten seeds and smoothed. Error bars show the standard error of the mean. }
    \label{fig:continual}
\end{figure}

We next test whether augmenting a reinforcement learner with the ISR module is sufficient to enable continual learning  \citep{thrun1998lifelong, hadsell2020embracing}. Figure~\ref{fig:continual} updates Figure~\ref{fig:challenge} and shows how a model with internalized reward (shown in red) performs when the social rewards from the caregiver are removed. The model with ISR continues to do the task at the same rate as one that continues receiving social rewards. Thus, for this context, the ISR model fully internalized the social rewards of the caregiver. This enables the agent to continue autonomously without dependence on the caregiver's social rewards to maintain its behavior.

The right panel of Figure~\ref{fig:continual} shows a test of generalization. During the socialization period, when the caregiver was present, the agent was trained with just one block in the environment for 6K episodes. We then tested how well the agent could learn directly from the ISR in environments with five blocks (another 6K episodes). Performance was evaluated by calculating the total reward that would have been obtained if the caregiver was present in a held-out set of 100 episodes, i.e., a proxy for how well the caregiver's values have been internalized and generalized. This is labeled Reward (OOD) on Figure~\ref{fig:continual}b. We compared ISR performance to a baseline (``frozen'') and an upper bound (``oracle''). The frozen baseline corresponds to testing a model right after the one-block socialization period on the five-block test without additional learning (all weights are frozen).

While we do see some generalization, the model with an ISR is able to continue learning on a five-block task and performs closer to the oracle, which learns directly from the ground truth caregiver's social rewards on the test environments.  See Figure~\ref{fig:continual} for statistical information. 

\begin{figure*}[!b]
    \centering
    \includegraphics[width=.8\linewidth,trim={0 4cm 0 8cm},clip]{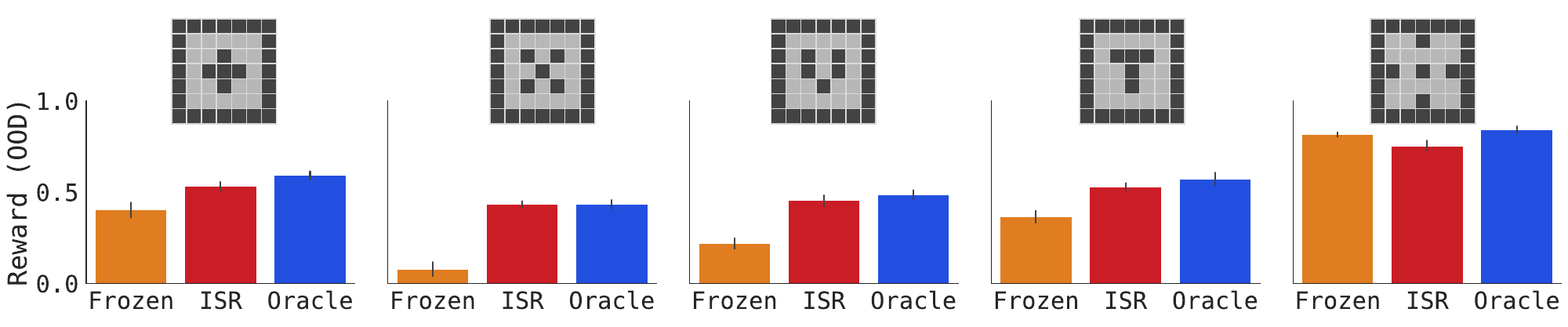}
    \caption{Out of distribution (OOD) generalization on custom environments. Agents were trained with only a single block and evaluated on their ability to generalize OOD to the above five block tasks. Starting location of the goal and agent was sampled randomly. The ISR significantly outperforms the frozen model ($p < 0.001$, $t=-3.66$, linear mixed-effect model with environment as a fixed effect) but did not significantly differ from the oracle ($p=.27$, $t=1.12$, linear mixed-effect model with environment as a fixed effect). Results are averaged over ten seeds and smoothed. Error bars are standard errors. }
    \label{fig:handpicked}
\end{figure*}

Finally, using the same paradigm as above, we looked at generalization performance on a few hand-chosen held-out grids shown in Figure~\ref{fig:handpicked}. In all but one case, the ISR outperforms the frozen baseline with performance approaching the oracle. In the one case where the agent with ISR did not outperform the baseline (``four rooms'' on the far right), performance was at ceiling for all models. 

\subsection{Internalization Failure: Reward Hacking}
\begin{figure}[!tb]
    \begin{minipage}[b]{0.65\linewidth}
    \includegraphics[width=\linewidth]{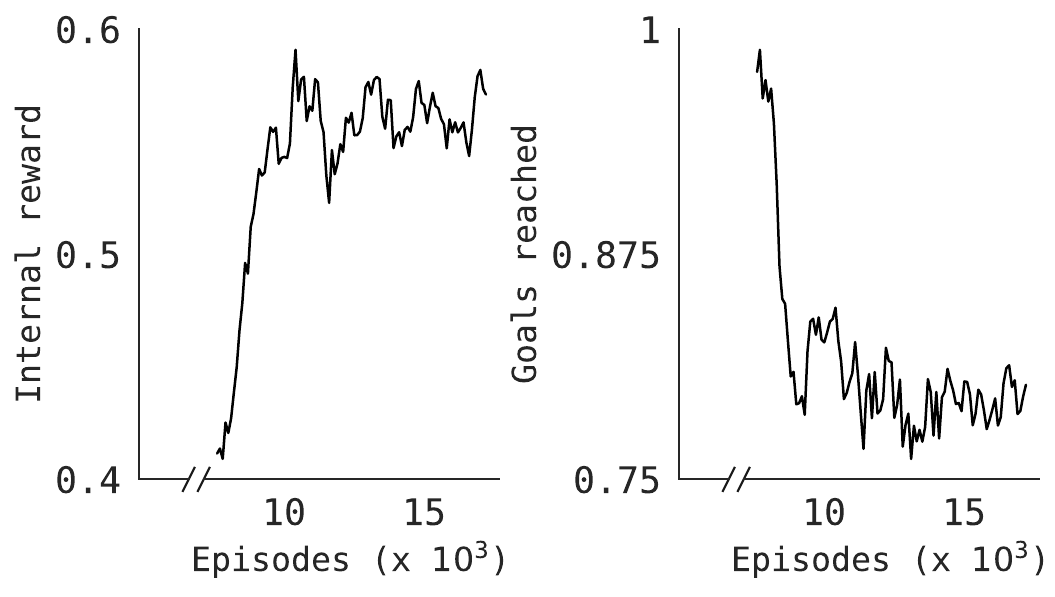}%
    \begin{tikzpicture}[overlay]
    \node[anchor=north east,xshift=+0cm,yshift=+5.5cm]
    {\includegraphics[width=2.3cm]{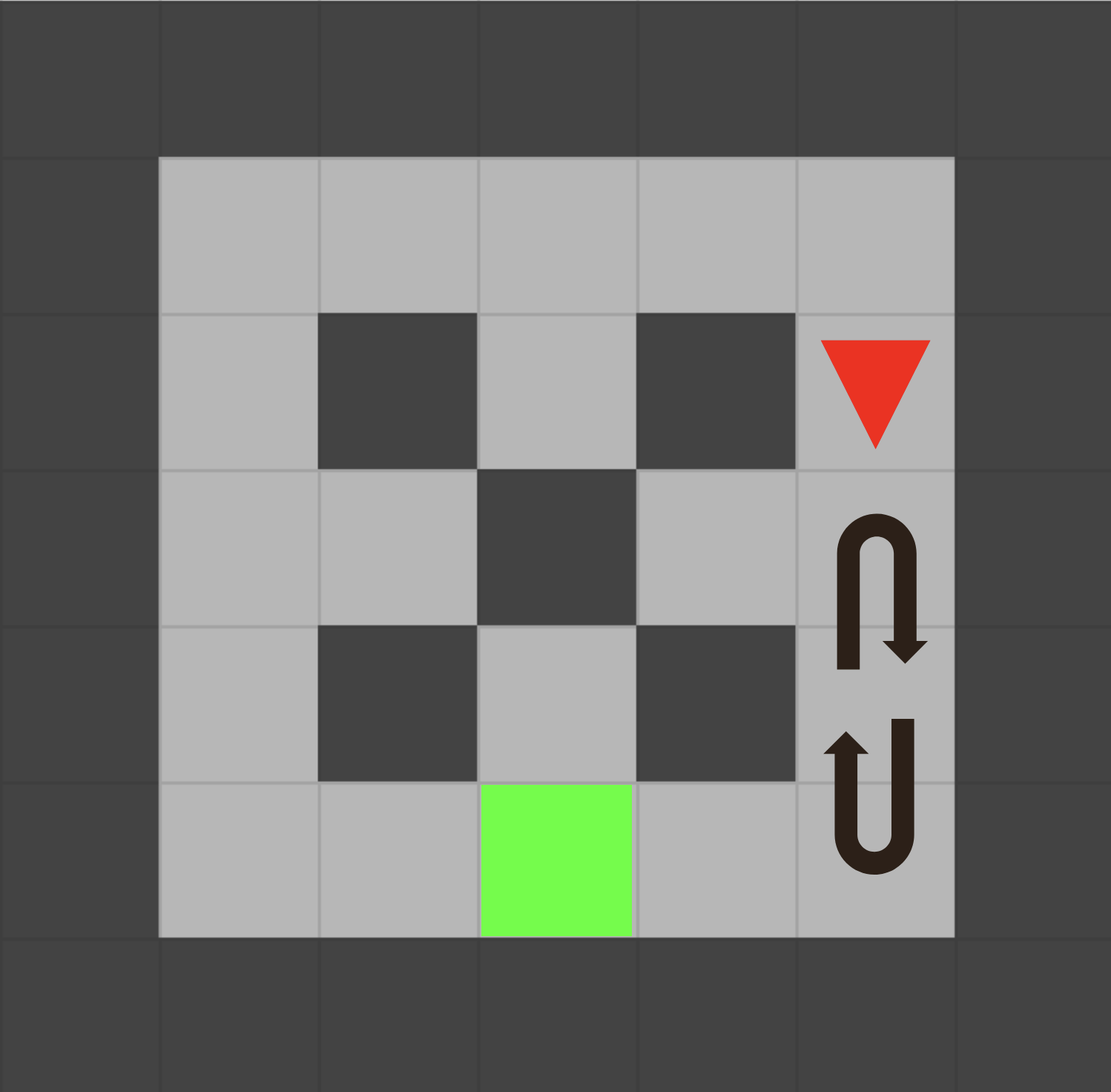}};
    \end{tikzpicture}
    \end{minipage}
    \begin{minipage}[b]{0.3\linewidth}
    \caption{Internalization failures. Reward hacking when the ISR model is undertrained. (left) After the caregiver leaves (break in the x-axis), internal rewards increase when learning from the ISR (right), but the number of goals reached declines. Inset shows an example of the loops that the agent learns in order to optimize internal reward.}
    \label{fig:hacking}
    \end{minipage}
\end{figure}

When value internalization is incomplete, problems can arise when the internalized rewards are prematurely optimized. Generalization failures in the ISR model will propagate into errors in the agent's policy during the autonomous period when the ISR model is the target for learning. To study this empirically, we undertrained the ISR model on 1/12th of the data as before. Figure~\ref{fig:hacking} shows that while the agent correctly optimizes its internal reward from the ISR model, it is less likely to reach the caregiver's goal. This failure can be considered an instance of reward hacking: the agent is optimizing for a proxy objective, the ISR, which diverges from the true objective, the caregiver's reward \citep{skalse2022defining}. The inset shows that the reward model is inconsistent, and the agent learns to loop around without reaching the green square.  

\subsection{Internalization of Prosocial Values} 
\begin{figure}[!b]
    \begin{subfigure}[b]{0.5\linewidth}
    \includegraphics[width=0.18\linewidth]{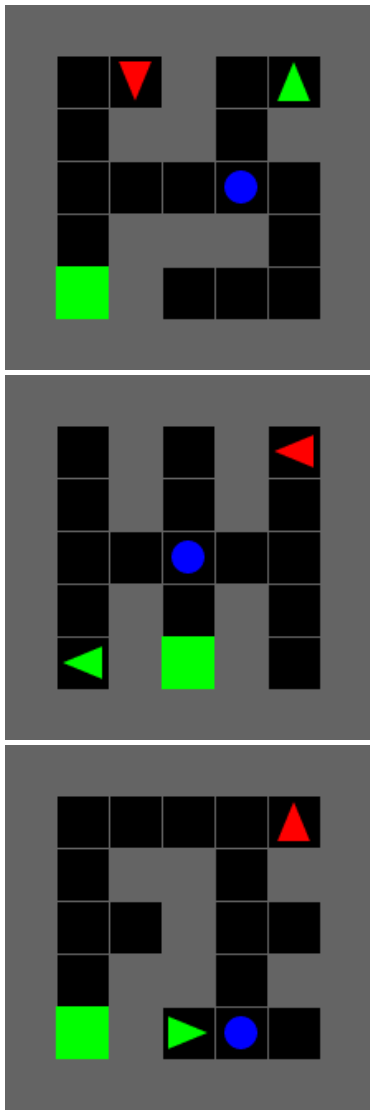}%
    \hfill
    \includegraphics[width=.78\linewidth]{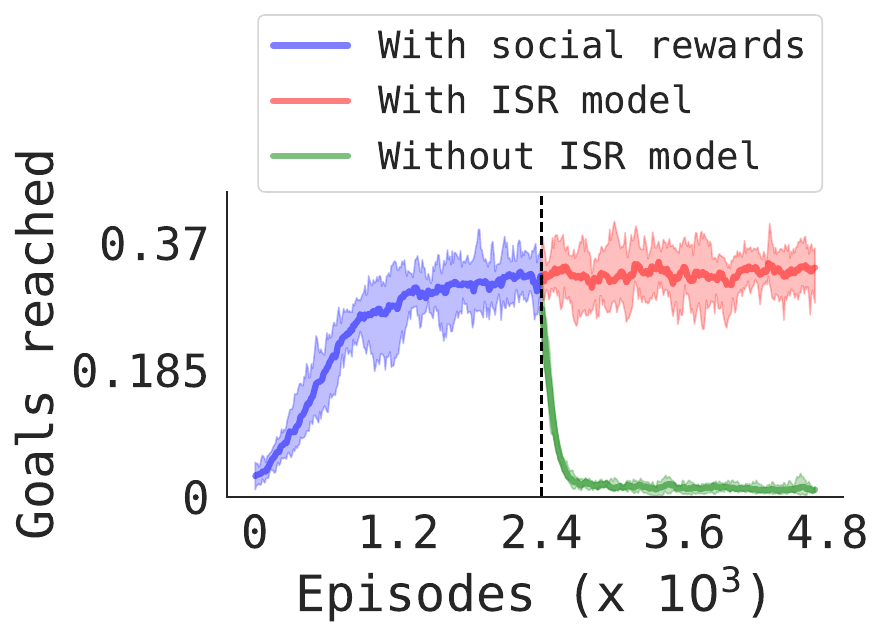}
    \end{subfigure}
    \begin{subfigure}[b]{0.5\linewidth}
    \end{subfigure}
    \begin{minipage}[b]{.5\linewidth}
    \caption{Prosocial value internalization. (left) Three procedurally generated environments where the red agent needs to pick up the blue boulder so the green agent can reach the green goal. (right) Agent's first learn to be prosocial with social rewards from the caregiver (blue). After 2.4K episodes, the caregiver is removed (vertical dotted line). Without the ISR model, the agent quickly unlearns the prosocial behavior (green). The ISR model prevents unlearning and the agent maintains a prosocial motivation (red). Results are averaged over ten seeds and smoothed. Bands show the min and max. 
    }
    \label{fig:prosocial}
    \end{minipage}
\end{figure}

Up to this point, our empirical investigation focused on a single agent operating alone in an environment. However, many of the most important culturally acquired values are interpersonal and relate to how we should treat others. We investigate this phenomenon in a procedurally generated set of two-player scenarios shown in Figure~\ref{fig:prosocial} inspired by \citet{ullman2009help}.  Instead of the red agent being socially rewarded when it reaches the green square, the caregiver rewards it when the green player reaches the green square. However, a blue boulder blocks the green arrow's path in each generated grid. The red agent has two additional actions ``pick up'' and ``drop'' which allow it to pick up and move the boulder, clearing the path. The state is also augmented to include a binary indicator of whether or not the player is carrying the boulder. If possible, the green agent always moves toward the goal using a depth-first search. If no path is found, it remains in place. 

To create solvable tasks procedurally, we generated 5x5 grids with seven blocks subject to the constraint that the remaining open tiles form a single connected component, which results in a tree structure. The boulder is placed in the location with the maximum degree in that tree, and the green player and green goal are placed on opposite components of the resulting disconnected graph at the tree's leaf nodes (endpoints). The red agent is placed at another leaf node. Thus, in each starting configuration, the green player's path is blocked by the boulder, and the only way for that player to reach the goal is if the red agent picks up and moves the boulder away. The caregiver gives a social reward to the red agent when the green player reaches the green square and otherwise gives no reward. Thus, we can study how a prosocial reward that is dependent on the behavior of another is internalized by the ISR module. 

Figure~\ref{fig:prosocial} shows the results from this experiment. Overall, we observe similar phenomena to those seen in previous experiments. During the socialization period, the agent learns the task. When the caregiver leaves, the agent without ISR unlearns the behavior. However, with ISR, the agent continues helping the other agent in new environments, having internalized the prosocial value. This is reminiscent of the human feeling of a ``warm glow'' when behaving altruistically \citep{andreoni1990impure}. 

\section{Discussion}
We develop a new computational cognitive model for studying how values can be socially acquired and maintained during learning. We proposed a process called value internalization, where, during a socialization period, a caregiver socially rewards a learning agent based on the correctness of their behavior. The learner models these rewards internally, and once the caregiver leaves and the learner must continue independently, the internal model of reward prevents unlearning the socially acquired behaviors and enables further learning and generalization. Together, these results shed light on some of the features and challenges of value acquisition. In the following, we discuss some implications that arise from this view and describe opportunities for future study on computational value internalization. 

Here, we only considered the simplest kind of social feedback, directly rewarding the desired outcome. However, human social feedback is far richer and often requires some computation on the side of the receiver to be interpreted correctly. For instance, when people teach with rewards and punishments, their actions have a communicative goal rather than just shaping a policy \citep{ho2017social, ho2019people}. In the prosocial environments shown in Figure~\ref{fig:prosocial}, rather than giving a reward when the green agent reaches the green square, it might be more natural to give a positive reward when the red agent picks up the boulder and moves it out of the way. Once the boulder has been moved, the red agent no longer has a role to play as a helper, so it might make sense to deliver the reward then. However, without additional inferential machinery, the agent will learn that moving boulders is the goal rather than seeing moving the boulder as a means to an end. A sophisticated learning agent should learn to disambiguate between approval of instrumentally valuable actions and intrinsically valuable actions when interpreting an approval signal provided by a caregiver. Other forms of social feedback, such as observation, demonstrations, language, or corrections, may need their own inferential machinery to distill into an ISR model \citep{colas2020language, jeon2020reward, kleiman2020downloading}. 

Why internalize social rewards instead of learning from scratch? Internalization is useful for any boundedly rational agent (including humans) that is unable to perform the (very) long-horizon planning required for survival in a complex world. Instead, a resource-rational strategy is to learn to intrinsically value the goals that have already been acquired by previous learners. These goals may end up somewhat decorrelated from the original environmental rewards, but may still lead to the acquisition of a wide variety of skills relevant to survival without the computational cost. Similar explanations have been offered for why it's adaptive for humans to internalize  social norms \citep{gintis2003hitchhiker}. 

Our computational approach to value internalization gives a novel view on a developmental question: when is an agent or organism ready to seek independence instead of further care? From the perspective of value internalization, the more time an agent spends with their caregiver, the more accurate their internal rewards will be (as we showed in Figure~\ref{fig:isr_train}). If we assume that the caregiver's social rewards transmit a culturally evolved set of values, then accurately representing those values will be of benefit to the learner \citep{henrich2015secret}. While a learner can only weakly estimate the benefit of a more accurate ISR model because of the uncertain future, an outer optimization loop of cultural evolution could at least estimate the average value of a given ISR accuracy \citep{sorg2010reward}. Let $B(n)$ be the benefit to a particular ISR and $n$ be the amount of social feedback that the ISR module was trained with. Furthermore, providing feedback is costly to the caregiver and may delay the productivity of the learner during independence. Let $C(n)$ be these costs, which are also a function of the amount of social rewards. Applying the logic of marginal utility, an agent is ready for independence when:
\begin{align*}
    \frac{\partial B}{\partial n} < \frac{\partial C}{\partial n}
\end{align*}
or when the marginal benefit of improving the ISR is less than the marginal cost of the next additional social reward. 

While this work used a deep neural network to model social reward, the framework we presented applies more generally to a wide range of representations. More structured models, such as hierarchical Bayesian models or probabilistic programs, may be better suited to capture people's inductive biases when learning what kinds of states and actions are likely to be rewarding \citep{kleiman2017learning}. These inductive biases give up some flexibility for greater sample efficiency. However, from an evolutionary perspective, flexibility might be highly valuable -- the range of possible cultural values cannot be easily anticipated (e.g., non-intuitive complex rituals) over the time span of biological evolution, and it may be worth spending more time and energy in a socialization phase to allow for a wider range of possible values \citep{piantadosi2016extraordinary}. For instance, what kind of ISR model could accurately recognize and reward values like curiosity and exploration? We hope to study this question empirically in future work. 

Value internalization may have implications for aligning artificial intelligence with human values. Today, large language models (LLMs) are made more helpful and ethical and less biased and harmful through a process called reinforcement learning from human feedback (RLHF) that shares a resemblance with the ISR model \citep{ouyang2022training}. Acting in a caregiver-like role, human annotators rate pairs of model outputs. Those ratings are used to train a reward model, which tunes the language model toward the preferences of the human annotators. In our work, we attempted to reverse engineer how human learners might internalize their caregiver's feedback, aligning their wants and desires to those of the previous generation. Now, we are faced with engineering these internalization mechanisms into AI agents in order to build safe, intelligent machines. 

\subsection{Source Code} \url{https://github.com/friedeggs/social-play}


\bibliography{aaai24}
\bibliographystyle{rlc}


\end{document}